\documentclass[10pt,twocolumn,twoside]{IEEEtran} \usepackage{graphicx} 
\usepackage{lipsum}
\usepackage{amsmath}
\usepackage{amsfonts}
\usepackage{amssymb}
\usepackage{bm}
\usepackage{tikz}
\usepackage{pgfplots}
\usepackage{hyperref}
\usepackage{comment}
\usetikzlibrary{calc,shapes.geometric}

\usepackage{amsthm}
\usepackage{xintexpr}
\usepackage{xinttools} 
\usepackage{tikz}
\usetikzlibrary{calc}
\usetikzlibrary{positioning}
\usetikzlibrary{shapes.multipart}
\usepackage{xcolor}
\usepackage{hyperref}
\usepackage[capitalise]{cleveref}

\tikzstyle{startstop} = [rectangle, rounded corners, minimum width=3cm, minimum height=1cm,text centered, draw=black, fill=red!30]
\tikzstyle{process} = [rectangle, minimum width=3cm, minimum height=1cm, text centered, draw=black, fill=orange!30]
\tikzstyle{decision} = [diamond, minimum width=3cm, minimum height=1cm, text centered, draw=black, fill=green!30]

\usepackage[citestyle=numeric-comp,sorting=none,url=true,doi=false,isbn=false]{biblatex}

\newcommand{\R}{\mathbb{R}}

\newcommand{\N}{\mathbb{N}}
\newcommand{\e}{\mathrm{e}}

\DeclareMathOperator*{\argmin}{arg\,min}

\title{Optimal Distributed Multi-Robot Communication-Aware Trajectory Planning using Alternating Direction Method of Multipliers}
\author{Jeppe Heini Mikkelsen, Roberto Galeazzi, and Matteo Fumagalli \\ Technical University of Denmark, Department of Electrical and Photonics Engineering}
\date{November 2023}

\begin{document}

\maketitle

\begin{abstract}
    This paper presents a distributed, optimal, communication-aware trajectory planning algorithm for multi-robot systems. Building on prior work, it addresses the multi-robot communication-aware trajectory planning problem using a general optimisation framework that imposes linear constraints on changes in robot positions to ensure communication performance and collision avoidance. In this paper, the optimisation problem is solved distributively by separating the communication performance constraint through an economic approach. Here, the current communication budget is distributed equally among the robots, and the robots are allowed to trade parts of their budgets with each other. The separated optimisation problem is then solved using the consensus alternating direction method of multipliers. The method was verified through simulation in an inspection task problem.
\end{abstract}

\section{Introduction}
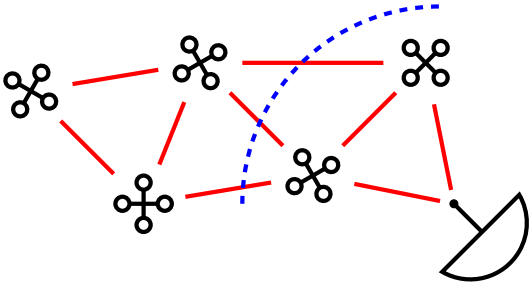
\begin{figure}[ht]
    \centering
    \begin{tikzpicture}[scale=0.75]
        \draw[ultra thick] (2.29289321881,-1.70710678119) arc (-120:30:1) -- cycle;
        \draw[ultra thick] (3,-1) -- (2.5,-0.5);
        \filldraw (2.5,-0.5) circle (2pt);
        \draw[ultra thick, red] (2.4510,-0.2549) -- (2.1471,1.2646);
        \draw[ultra thick, red] (0.7354,-0.1471) -- (2.2549,-0.4510);
        \draw[ultra thick, red] (0.5303,0.5303) -- (1.4697,1.4697);
        \draw[ultra thick, red] (-1.2500,2.0000) -- (1.2500,2.0000);
        \draw[ultra thick, red] (-2.2785,1.3036) -- (-2.7215,0.1964);
        \draw[ultra thick, red] (-0.7398,-0.1233) -- (-2.2602,-0.3767);
        \draw[ultra thick, red] (-1.4697,1.4697) -- (-0.5303,0.5303);
        \draw[ultra thick, red] (-4.2602,1.6233) -- (-2.7398,1.8767);
        \draw[ultra thick, red] (-4.4697,0.9697) -- (-3.5303,0.0303);

        \foreach \x/\y/\scale/\rotate in {0/0/0.25/30, -2/2/0.25/30, 2/2/0.25/-45, -3/-0.5/0.25/90, -5/1.5/0.25/60}
        {
            \begin{scope}[shift={(\x,\y)}, scale=\scale, rotate=\rotate]
                \draw[ultra thick] (-1,0) -- (1,0);
                \draw[ultra thick] (0,-1) -- (0,1);
                \draw[ultra thick] (-1.5,0) circle (0.5);
                \draw[ultra thick] (1.5,0) circle (0.5);
                \draw[ultra thick] (0,-1.5) circle (0.5);
                \draw[ultra thick] (0,1.5) circle (0.5);
            \end{scope}
        }

        \draw[ultra thick, blue, dashed] (-1.25,-0.5) arc (-180:-270:3.5);
    \end{tikzpicture}
    \caption{An example of an ad-hoc network where five drones communicate wirelessly with each other and a ground station antenna. Three drones outside the antenna's range (blue dashed line) rely on each other to relay their information \cite{Mikkelsen2024OptimalValue}.}
    \label{fig:network}
\end{figure}

Many multi-robot systems rely on wireless communication to achieve coordination in planning and sensing scenarios. Wireless communication have several limitations, primarily due to path loss, shadowing, and multi-path fading, which affect transmission distance and data throughput. These issues create a dependency between the position of the robots and their communication capability, necessitating the use of communication-aware trajectory planners (CaTP) \cite{Licea2022WhenTutorial}. Broadly speaking, wireless communication can be broken into two categories: infrastructure based communication, and ad-hoc communication. In infrastructure based communication data is routed through a fixed network infrastructure that each robot have to connect to directly. In ad-hoc networks robots are able to route data, effectively extending the range of the network and increasing its robustness, see \cref{fig:network}. However, since the robots rely on each other to route data, the communication network performance becomes dependent on inter-robot distances, further increasing the complexity of coordination and necessitating multi-robot communication-aware trajectory planners (MR-CaTP). In \cite{Mikkelsen2024OptimalValue} it was shown that the Fiedler value can be used as a communication performance metric in ad-hoc networks. The Fiedler value is the second smallest eigenvalue of the communication networks graph Laplacian, and is strictly positive as long as the communication network remains connected. There is rich literature on MR-CaTP using the Fiedler value. A common method is to move the robots in order to maximise the Fiedler value, and thereby the connectivity \cite{DeGennaro2006DecentralizedSystems,Kim2006OnLaplacian,Yang2008DecentralizedNetworks,EthanStump2008ConnectivityTeams}. However, the robots might have alternative goals that conflict with maximising the Fiedler value where, e.g., some of the robots are assigned to inspect points of interest (POI), and might move faster towards the POIs than the remaining robots can ensure that the Fiedler value stays strictly positive while moving to maximise it. In \cite{Mikkelsen2024OptimalValue} a method for ensuring communication performance by applying a linear inequality constraint on the trajectories of the robots was introduced. The method uses a first order Taylor expansion to estimate the change in Fiedler value from the change in position of the robots, which is then used to constrain the predicted Fiedler value in a receding horizon optimisation framework. The optimisation problem is solved at a central location, either a base-station or a robot assigned as a leader. This presents a risk to the multi-robot system, as the central location represents a single point of failure (SPOF). In this paper, the work in \cite{Mikkelsen2024OptimalValue} is extended to solve the optimisation problem distributively. The method in this paper draws inspiration from the consensus alternating direction method of multipliers (C-ADMM) \cite{Boyd2010DistributedMultipliers} and the separable optimisation variable ADMM (SOVA) method \cite{Shorinwa2020ScalableNetworks}. In \cite{Boyd2010DistributedMultipliers}, it is shown that when the cost function is a sum of sub-cost functions, the optimisation problem can be decomposed into several independently solvable problems, using a dual-ascent approach to ensure consensus. However, complexity increases as the number of robots grows since each sub-cost function depends on the full variable set. In \cite{Shorinwa2020ScalableNetworks}, it is demonstrated that if each sub-cost function depends only on a subset of the problem variables, the optimisation problem can be further decomposed into simpler independently solvable sub-problems, also ensuring consensus using dual ascent. For an in-depth review of distributed optimisation for multi-robot systems, see \cite{HalstedASystems}. In \cite{Zavlanos2013NetworkNetworks}, the authors present a distributed optimisation method for computing optimal routing variables in a network, used alongside a potential field method for ensuring network integrity. Similarly, in \cite{Kantaros2016GlobalEnvironments} optimal routing variables are computed distributively and integrated with a planner to grow a communication tree for servicing tasks.

The main contribution of this paper is a novel method for making an approximate separation of a non-separable linear inequality constraint based on an economic interpretation, allowing the MR-CaTP problem to be solved distributively using a dual-ascent approach. The presented method is capable of imposing hard constraints on the Fiedler value of the communication network. Existing methods based on the Fiedler value are either fully centralised or rely on moving the robots according to the gradient of the Fiedler value in order to maximise it. This is limiting since it cannot guarantee that the Fiedler value remains above some lower bound, and it might conflict with higher level goals that the multi-robot system has to achieve. Due to the generality of the proposed method, we postulate that this method can also be applied to a broader set of multi-agent optimisation problems with shared resource constraints. Lastly, the distributed optimisation algorithm presented in this paper relies on the robots executing certain processes synchronously. This requires that they simultaneously transition between processes when all robots are ready. To achieve this a leaderless method for determining when the processes in the MR-CaTP algorithm have converged and when to transition, based on a distributed consensus algorithm, is presented.

Throughout the paper, we adopt the following notation style: italic symbols $x/X$ denote scalars; bold italic symbols \(\bm{x}/\bm{X}\) denote vectors; bold non-italic symbols \(\mathbf{x}/\mathbf{X}\) denote matrices; and calligraphic symbols \(\mathcal{X}\) denote sets, and $|\mathcal{X}|$ denotes the cardinality of the set $\mathcal{X}$. Notations \(\R_{\geq0}\) and \(\R_{>0}\) denote non-negative and strictly positive real numbers, respectively. Subscripts $\mathbf{X_{ij}}$ denote the $i$\textsuperscript{th} and $j$\textsuperscript{th} index of the matrix. Square brackets $[\cdot,\cdot]$ indicate column concatenation and parentheses $(\cdot,\cdot)$ indicate row concatenation. All vectors are column vectors.

\section{Robot and Communication Network Model}
Consider a swarm of $N$ robots with indices denoted by $\mathcal{V}$, with the position of robot $i$ at time $t$ given by $\bm{p_i}(t) \in \R^n$. Each robot is holonomic and occupies a region defined by a hyper-sphere centred at $\bm{p_i}(t)$ with radius $r_i$. The movement of the robots occurs in discrete time steps $\Delta t$:
\begin{equation}\label{eq:discrete_model}
    \bm{p_i^{k+1}} = \bm{p_i^k} + \bm{u_i^k}, \ \forall \ i \in \mathcal{V},
\end{equation}
where $k$ denotes the time step, $\bm{p_i^k} = \bm{p_i}(k\Delta t)$, and $\bm{u_i^k}$ is the positional change between steps, referred to as the robot's input at time step $k$. The robots communicate via wireless links, forming a time-varying undirected graph $\mathcal{G}(t) = (\mathcal{V}, \mathcal{E}(t))$, with $\mathcal{E}(t)$ representing the communication links. Like in \cite{Mikkelsen2024OptimalValue}, each link has an associated weight $w_{ij} \in [0,1]$, which represents the probability of successful packet transmission from robot $i$ to robot $j$. This weight is modeled as a function of the distance between the robots using the logistic function \cite{Zavlanos2011Graph-theoreticNetworks}:
\begin{equation}\label{eq:prr}
w_{ij}(t) =
\begin{cases}
    \dfrac{\e^{-\alpha(d_{ij}(t) - d_{50})}}{1 + \e^{-\alpha(d_{ij}(t) - d_{50})}} & \text{if } (i,j) \in \mathcal{E}, \\
    0 & \text{otherwise},
\end{cases}
\end{equation}
\begin{equation}
    d_{ij}(t) = \left|\bm{p_i}(t) - \bm{p_j}(t)\right|_2,
\end{equation}
where $d_{50}$ is the distance at which the link quality is 50\%, and $\alpha$ is four times the attenuation rate at $d_{50}$.

\begin{figure}[ht]
    \centering
    \begin{tikzpicture}
        \begin{axis}[
            domain=0:4, 
            samples=500,
            axis lines=middle, 
            grid=both, 
            xmin=-0.25, 
            xmax=4.2, 
            ymin=-0.25, 
            ymax=1.2, 
            xlabel={$d_{ij}$}, 
            ylabel={$w_{ij}$},
            xtick=\empty,
            width=\linewidth,
            height=0.5\linewidth
        ]
            \addplot[red, thick] {1 - 1/(1 + exp(-4*(x - 2)))};
            \addplot[green, thick] {1 - 1/(1 + exp(-8*(x - 2)))};
            \addplot[blue, thick] {1 - 1/(1 + exp(-16*(x - 2)))};
            
            \draw[gray] (axis cs:2,0) -- (axis cs:2,1.2) node [anchor=north,color=black,pos=0] {$d_{50}$};
        \end{axis}
    \end{tikzpicture}
    \caption{Link quality with increasing values of $\alpha$, in order of red, green, and blue \cite{Mikkelsen2024OptimalValue}.}
    \label{fig:packet_arrival}
\end{figure}
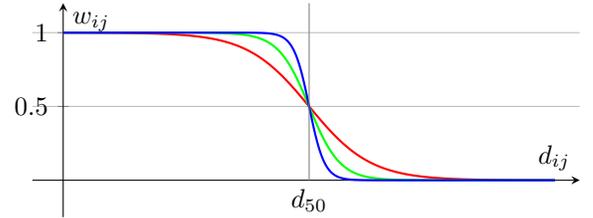

\section{Communication Performance Metric}
In \cite{Mikkelsen2024OptimalValue}, it was found that the Fiedler value can be used as a communication performance metric. For a graph $\mathcal{G}(t)$, the adjacency matrix $\mathbf{A}(t) = (a_{ij}(t)) \in [0,1]^{N\times N}$ is a hollow symmetric matrix, and the degree matrix $\mathbf{D}(t) = (d_{ij}(t)) \in [0,N-1]^{N\times N}$ is a diagonal matrix defined as:
\begin{gather}
    a_{ij}(t) = w_{ij}(t) \ \forall \ i \neq j \in \mathcal{V}, \\ 
    d_{ii}(t) = \sum_{j\in\mathcal{V}} a_{ij}(t) \ \forall \ i \in \mathcal{V}.
\end{gather}
The graph Laplacian is:
\begin{equation}
    \mathbf{L}(t) = \mathbf{D}(t) - \mathbf{A}(t),
\end{equation}
with eigenvalues $\lambda_1(t) \leq \lambda_2(t) \leq \dots \leq \lambda_N(t)$ and eigenvectors $\bm{v_1}(t), \bm{v_2}(t), \dots, \bm{v_N}(t)$. The eigenvalues are bounded between $0$ and $N$, with $\lambda_1(t) = 0$ and $\bm{v_1}(t) = \bm{1}$ for all $t$. The second smallest eigenvalue, $\lambda_2(t)$, is the Fiedler value. If $\lambda_2(t) = 0$ the graph is disjoint, indicating loss of communication at time $t$; if $\lambda_2(t) = N$ the graph is complete, indicating perfect communication at time $t$. To maintain connectivity, $\lambda_2(t)$ must stay above a lower bound $\underline{\lambda}_2 > 0$. For algorithmic purposes, the graph properties and Fiedler value are computed at discrete time steps:
\begin{equation}
    \begin{aligned}
        &\mathbf{A^k} = \mathbf{A}(k\Delta t), &\mathbf{D^k} = \mathbf{D}(k\Delta t), \\
        &\mathbf{L^k} = \mathbf{L}(k\Delta t), &\lambda_2^k = \lambda_2(k\Delta t).
    \end{aligned}
\end{equation}
\section{Optimisation Problem}\label{sec:optimisation_problem}
Given the communication performance metric and its lower bound, the MR-CaTP problem can be formulated as the following optimisation problem:
\begin{align}
    \bm{u^{k*}} = \argmin_{\bm{u^k}} & \quad \sum_{i\in\mathcal{V}} f_i(\bm{u_i^k}) \\
    \text{s.t.} \quad & \bm{p_i^{k+1}} = \bm{p_i^k} + \bm{u_i^k} \ \forall \ i \in \mathcal{V}, \label{eq:kin_const}\\
    & ||\bm{u_i^k}||_p \leq \bm{\overline{u}} \ \forall \ i \in \mathcal{V}, \label{eq:vel_const}\\
    & \lambda_2^{k+1} \geq \underline{\lambda}_2, \label{eq:com_const}\\
    & ||\bm{p_i^{k+1}} - \bm{p_j^{k+1}}||_2 \geq r_i + r_j + \varepsilon \dots \\ 
    &\forall \ i \neq j \in \mathcal{V}\times\mathcal{V}, 
    \label{eq:col_const}
\end{align}
where $\bm{u^k} = (\bm{u_1^k},\dots,\bm{u_N^k})$ is the concatenated robot input, $f_i(\cdot)$ is the cost function for robot $i$'s input, \eqref{eq:kin_const} represents the robot kinematics, \eqref{eq:vel_const} is an input norm constraint, \eqref{eq:com_const} is a constraint on the future Fiedler value, and \eqref{eq:col_const} enforces collision avoidance with minimum clearance $\varepsilon$. The cost function is separable, however the constraints in \eqref{eq:com_const} and \eqref{eq:col_const} are not, preventing the optimal solution from being computed locally on each robot. The following sections present a method to make the optimisation problem separable.

\section{Separable Fiedler Value Approximation}
Typically, imposing a constraint on a matrix eigenvalue requires a semi-definite constraint, as shown in \cite{Derenick2010AEnvironments}. This is not separable because the contribution of each robot to the change in Fiedler value is not explicitly calculated. In \cite{Mikkelsen2024OptimalValue}, a first-order Taylor expansion approximating the change in Fiedler value from the change in robot position is used to predict the Fiedler value in the communication constraint in \eqref{eq:com_const} as
\begin{equation}\label{eq:com_const_approx}
    \hat{\lambda}_2^{k+1} = \lambda_2^k + \sum_{i\in\mathcal{V}} \bm{m_i^{k\top}}\bm{u_i^k} \geq \underline{\lambda}_2,
\end{equation}
see \cref{sec:fiedler_derivative} for the derivation. The communication constraint in \eqref{eq:com_const_approx} can be rearranged into:
\begin{equation}
    \underbrace{- \sum_{i\in\mathcal{V}} \bm{m_i^{k\top}}\bm{u_i^k}}_{\text{expenditure}} \leq \underbrace{\lambda_2^k - \underline{\lambda}_2}_{\text{budget}},
\end{equation}
where $\lambda_2^k - \underline{\lambda}_2$ and $-\bm{m_i^{k\top}}\bm{u_i^k}$ represents the communication budget and the expenditure of each robot respectively. Distributing the budget equally among the robots results in $N$ separate constraints:
\begin{equation}
    - \bm{m_i^{k\top}}\bm{u_i^k} \leq \frac{1}{N}(\lambda_2^k - \underline{\lambda}_2) \ \forall \ i \in \mathcal{V}.
\end{equation}
To address the limitation of equal distribution, robots are allowed to exchange part of their budget with their neighbours:
\begin{equation}
    - \bm{m_i^{k\top}}\bm{u_i^k} \leq \frac{1}{N}(\lambda_2^k - \underline{\lambda}_2) + \sum_{j \in \mathcal{N}_i} t_{ij}^k  \ \forall \ i \in \mathcal{V},
\end{equation}
where $t_{ij}^k$ is the trading variable between robot $i$ and $j$. To ensure agreement on budget exchange, the following consensus constraint on the trading variables is imposed:
\begin{equation}
    t_{ij}^k = - t_{ji}^k \ \forall \ \{i,j\} \in \mathcal{V}\times\mathcal{N}_i.
\end{equation}
This constraint can be separated using a dual-ascent optimisation approach, elaborated in \cref{sec:opt_prob_appr}. Lastly, the trading variables are gathered into a vector $\bm{t_i^k} \in \R^{|\mathcal{N}_i|}$ and moved to the left-hand side:
\begin{equation}\label{eq:dist_com_const_approx}
    - \bm{m_i^{k\top}}\bm{u_i^k} - \bm{1_{|\mathcal{N}_i|\times 1}^\top}\bm{t_i^k} \leq \frac{1}{N}(\lambda_2^k - \underline{\lambda}_2).
\end{equation}
Thereby, the single communication constraint is separated into $N$ communication constraints, one for each robot.

\section{Collision Avoidance}
Like in \cite{Zhou2017FastCells}, the robots are constrained to lie within disjoint sets to avoid collisions. This is done using a separating hyperplane between each robot pair. Thus, the feasible position set for robot $i$ becomes
\begin{equation}
    \mathcal{P}_i^k = \{p\in\R^n \ | \ \bm{c_{ij}^{k\top}}\bm{p} \leq d_{ij}^k \ \forall \ j \in \mathcal{N}_i\},
\end{equation}
where
\begin{equation}
    \bm{c_{ij}^k} = \frac{\bm{p_j^k} - \bm{p_i^k}}{||\bm{p_j^k} - \bm{p_i^k}||_2}, \quad d_{ij}^k = \frac{1}{2}\bm{c_{ij}^{k\top}}(\bm{p_i^k} + \bm{p_j^k}) - (r_i + \varepsilon/2), 
\end{equation}
with $r_i$ being the radius of robot $i$ and $\varepsilon$ being the minimum clearance between the robots. The hyperplane is buffered by half the clearance to ensure that the robots keep a safe distance from each other. The constraint for each neighbour can be stacked together to make the constraint
\begin{equation}\label{eq:col_const_approx}
    \mathbf{C_i^k}\bm{p_i^k} \leq \bm{d_i^k}.
\end{equation}
where
\begin{gather}
    \mathbf{C_i^k} = [\bm{c_{i\mathcal{N}_{i}[1]}^k},\dots,\bm{c_{i\mathcal{N}_{i}[|\mathcal{N}_i|]}^k}]^\top, \\
    \bm{d_i^k} = (d_{i\mathcal{N}_{i}[1]}^k,\dots,d_{i\mathcal{N}_{i}[|\mathcal{N}_i|]}^k).
\end{gather}
\section{$M$-Step Prediction}
To avoid greedy solutions, the optimisation problem is augmented into an $M$-step receding horizon problem, where the inputs, communication expenditure, and trading variables are predicted $M$-steps into the future, the first position change is applied, and the process is repeated indefinitely or until the algorithm is terminated. The $M$ future positions are predicted as
\begin{equation}\label{eq:pos_pred}
    \bm{P_i^{k+1}} = \bm{1_{M\times1}}\otimes\bm{p_i^{k}} + \mathbf{B}\bm{U_i^k}
\end{equation}
where
\begin{equation}
    \mathbf{B} = \mathbf{L_M} \otimes \mathbf{I_{n}}
\end{equation}
where $\mathbf{L_M}$ is an $M\times M$ lower triangular matrix of ones, $\mathbf{I_{n}}$ is a $n\times n$ identity matrix, $\otimes$ is the Kronecker product operator, and
\begin{equation}
    \bm{P_i^{k+1}} = (\bm{p_i^{k+1}},\bm{p_i^{k+2}},\dots,\bm{p_i^{k+M}}),
\end{equation}
\begin{equation}
    \bm{U_i^k} = (\bm{u_i^k},\bm{u_i^{k+1}},\dots,\bm{u_i^{k+M-1}}).
\end{equation}
The distributed communication constraint in \eqref{eq:dist_com_const_approx} can be turned into an $M$-step constraint as
\begin{equation}\label{eq:rh_dist_com_const_approx}
    -\mathbf{M_i^k}\bm{U_i^k} - \mathbf{F_i^k}\bm{t_i^k} \leq \bm{1_{M\times1}}\frac{1}{N}(\lambda_2^k - \underline{\lambda}_2),
\end{equation}
where
\begin{equation}
    \mathbf{M_i^k} = \mathbf{L_M}\otimes\bm{m_i^{k\top}}, \quad \mathbf{F_i^k} = \mathbf{1_{M\bm{\times|\mathcal{N}_i|}}}.
\end{equation}
To reduce the computational complexity of solving the optimisation problem, the trading variables are held constant throughout the prediction window. Using the $M$-step prediction model in \eqref{eq:pos_pred}, the approximate collision avoidance constraint in \eqref{eq:col_const_approx} can be turned into an $M$-step constraint as
\begin{equation}\label{eq:rh_col_const_approx}
    \mathbf{\Tilde{C}_i^k}\bm{U_i^k} \leq \bm{\Tilde{d}_i^k},
\end{equation}
where
\begin{equation}
    \mathbf{\Tilde{C}_i^k} = (\mathbf{I_M}\otimes \mathbf{C_i^k})\mathbf{B}, \quad \bm{\Tilde{d}_i^k} = \bm{1_{M\times1}}\otimes(\bm{d_i^k} - \mathbf{C_i^k}\bm{p_i^k}).
\end{equation}

\section{Fiedler Value and Vector Estimation}
Predicting the Fiedler value in \eqref{eq:rh_dist_com_const_approx} requires the current Fiedler value and vector to be known by all robots. Since these are calculated from the adjacency matrix, this requires global knowledge of all robot positions. To mitigate this, the Fiedler value and vector is replaced with estimates $\hat{\lambda}_2^k$ and $\bm{\hat{v}_2^k}$ respectively. To estimate the Fiedler value and vector, the adjacency matrix is estimated as $\mathbf{\hat{A}_i^k} = (\hat{a}_{jl,i}^k) \in [0,1]^{N\times N}$ on each robot using max consensus \cite{Iutzeler2012AnalysisChannels}. The entry corresponding to the communication neighbours of each robot is found as the expected packet reception rate using \eqref{eq:prr}
\begin{equation}
    \hat{a}_{ij,i}^k = \hat{a}_{ji,i}^k = w_{ij}^k \ \forall \ j \in \mathcal{N}_i.
\end{equation}
Robots can also aid in calculating the expected packet reception rate between neighbours on their behalf
\begin{equation}
    \hat{a}_{jl,i}^k = w_{jl}^k \ \forall \ j\neq l \in \mathcal{N}_i.
\end{equation}
This allows the robots to know the expected packet reception rate with their 1-hop neighbours as well, aiding them in planning trajectories where a direct connection with their 1-hop neighbours is established if needed. For the remaining off-diagonal entries the estimate is updated using a max consensus
\begin{equation}
        \hat{a}_{jl,i}^k \leftarrow \max_{h\in\mathcal{N}_i} \hat{a}_{jl,h}^{k-1} \ \forall \ j \notin \{i,\mathcal{N}_i\}, l \notin \{i,\mathcal{N}_i\},
\end{equation}
and the diagonal entries are kept at zero
\begin{equation}
    \hat{a}_{jj,i}^k = 0 \ \forall \ j \in \mathcal{V}.
\end{equation}
Once the estimates of the adjacency matrix entries have converged for all robots, the Fiedler value and vector is estimated locally on each robot by solving the eigenvalue equation
\begin{equation}
    \mathbf{\hat{A}_i^k}\bm{\hat{v}_{2,i}^k} = \hat{\lambda}_{2,i}^k\bm{\hat{v}_{2,i}^k}.
\end{equation}

\section{Optimisation Problem Approximation}\label{sec:opt_prob_appr}
Having derived an $M$-step approximate communication constraint in \eqref{eq:rh_dist_com_const_approx}, collision avoidance constraint in \eqref{eq:rh_col_const_approx}, and Fiedler value and vector estimation, the original MR-CaTP optimisation problem can be approximated as $N$ separate optimisation problems
\begin{align}
    \bm{U_i^{k*}} = & \argmin_{\bm{U^k}} \quad \sum_{m=0}^{M-1} f_i(\bm{u_i^{k+m}}) \\
    &s.t. \quad \\
    &\bm{P_i^{k+1}} = \bm{1_{M\times1}}\otimes\bm{p_i^{k}} + \mathbf{B}\bm{U_i^k}, \\
    & ||\bm{u_i^{k+m}}||_p \leq \bm{\overline{u}} \ \forall \ m \in \{0,\dots,M-1\}, \\
    &  -\mathbf{M_i^k}\bm{U_i^k} - \mathbf{F_i^k}\bm{t_i^k} \leq \bm{1_{M\times1}}\frac{1}{N}(\hat{\lambda}_{2,i}^k - \underline{\lambda}_2), \\
    & \mathbf{\Tilde{C}_i^k}\bm{U_i^k} \leq \bm{\Tilde{d}_i^k}, \\
    & \bm{t_{i}^{k}} = -\bm{t_{\mathcal{N}_i}^{k}}. \label{eq:consensus_const}
\end{align}
where $\bm{t_{\mathcal{N}_i}^k} \in \R^{M|\mathcal{N}_i|}$ are the trading variables between robot $i$ and its neighbours $\mathcal{N}_i$. The constraint in \eqref{eq:consensus_const} does not seem separable since it involves decision variables related to robot $i$'s neighbours. However, using a dual-ascent approach, this constraint can also be made separable \cite{Boyd2010DistributedMultipliers}. The constraint in \eqref{eq:consensus_const} is relaxed and applied as a cost term in the optimisation problem
\begin{align}
    &\begin{aligned}
    \{\bm{U_i^{k*}},\bm{t_i^{k*}}\} = & \argmin_{\bm{U_i^{k}},\bm{t_i^{k}}} \quad \sum_{m=0}^{M-1} f_i(\bm{u_i^{k+m}}) + \dots \\ & \bm{\mu_i^\top}(\bm{t_i^k} + \bm{t_{\mathcal{N}_i}^{k-1*}}) + \frac{\rho}{2}||\bm{t_i^k} + \dots \\ & \bm{t_{\mathcal{N}_i}^{k-1*}}||_2,
    \end{aligned} \label{eq:dist_opt_prob}\\
    &s.t. \quad \\
    & \qquad \bm{P_i^{k+1}} = \bm{1_{M\times1}}\otimes\bm{p_i^{k}} + \mathbf{B}\bm{U_i^k}, \\
    & \qquad ||\bm{u_i^{k+m}}||_p \leq \bm{\overline{u}} \ \forall \ m \in \{0,\dots,M-1\}, \\
    & \qquad -\mathbf{M_i^k}\bm{U_i^k} - \mathbf{F_i^k}\bm{t_i^k} \leq \bm{1_{M\times1}}\frac{1}{N}(\hat{\lambda}_{2,i}^k - \underline{\lambda}_2), \\
    & \qquad \mathbf{\Tilde{C}_i^k}\bm{U_i^k} \leq \bm{\Tilde{d}_i^k},
\end{align}
where $\bm{t_{\mathcal{N}_i}^{k-1*}}$ are the optimal trading variables from robot $i$'s neighbours from the previous time step and $\bm{\mu_i^k} = (\mu_{j,i}^k) \in \R^{|\mathcal{N}_i|\times 1}$ is a penalty multiplier associated with each trading variable. At the start of every optimisation period, the trading variables and penalty multipliers are initialised to zero for all robots. The robots exchange their previous trading variables and calculate their optimal trajectories and current trading variables by solution of \eqref{eq:dist_opt_prob}. They then update the penalty multipliers according to
\begin{equation}
    \bm{\mu_i^{k}} = \bm{\mu_i^{k-1}} + \rho(\bm{t_i^{k*}} + \bm{t_{\mathcal{N}_i}^{k-1*}}).
\end{equation} 
The robots repeatedly exchange their previous trading variables, solve their respective optimisation problems, and update their penalty multipliers. As the penalty multiplier increases the trading variables are driven towards consensus. Once the optimisation has converged for all robots, the trading variables are averaged for each neighbour to ensure complete agreement
\begin{equation}
    \bm{t_i^{k*}} = \frac{1}{2}(\bm{t_i^{k-1*}} - \bm{t_{\mathcal{N}_i}^{k-1*}}),
\end{equation}
the optimisation is performed one more time without trading
\begin{align}
    \bm{U_i^{k*}} = & \argmin_{\bm{U^k}} \quad \sum_{m=0}^{M-1} f_i(\bm{u_i^{k+m}}) \\
    &s.t. \quad \\
    &\bm{P_i^{k+1}} = \bm{1_{M\times1}}\otimes\bm{p_i^{k}} + \mathbf{B}\bm{U_i^k}, \\
    & ||\bm{u_i^{k+m}}||_p \leq \bm{\overline{u}} \ \forall \ m \in \{0,\dots,M-1\}, \\
    &  -\mathbf{M_i^k}\bm{U_i^k} \leq \bm{1_{M\times1}}\frac{1}{N}(\hat{\lambda}_{2,i}^k - \underline{\lambda}_2) + \mathbf{F_i^k}\bm{t_i^{k*}}, \\
    & \mathbf{\Tilde{C}_i^k}\bm{U_i^k} \leq \bm{\Tilde{d}_i^k}, \\
\end{align}
and the position reference of the robots is updated to
\begin{equation}
    \bm{p_{ref,i}^{k+1}} = \bm{p_i^k} + \bm{u_i^{k*}}, \ \forall \ i\in\mathcal{V}.
\end{equation}
\section{Convergence Consensus}


Since the Fiedler estimation and the optimisation has to happen sequentially and synchronously across all robots, the robots need to reach consensus on when convergence has been reached and when to switch process. To do this, each robot stores a boolean vector $\bm{b_i^k} = (b_{ij}^k) \in \{True,False\}^N$, an integer vector $\bm{d_i^k} = (d_{ij}^k) \in \N^N$ and an integer value $s_i^k \in \N$. The boolean vector stores whether each robot in the formation has converged, the integer vector stores the distance in the graph from robot $i$ to each of the robots, and the integer value stores at which time index the robots should transition. The vectors are initialised as follows
\begin{gather}
    b_{ij}^0 = False \ \forall \ (i,j) \in \mathcal{V}\times\mathcal{V}, \\
    d_{ij}^0 = \infty \ \forall \ i \in \mathcal{V}, \ j \neq i \in \mathcal{V}, \\
    d_{ii}^k = 0 \ \forall \ i \in \mathcal{V}, \ k \in \N, \\
    s_i^0 = \infty \ \forall \ i \in \mathcal{V}.
\end{gather}
If robot $i$ has converged at time-step $k$, it updates the entry at the $i$\textsuperscript{th} index of its boolean vector to $True$
\begin{equation}
    b_{ii}^k = 
    \begin{cases}
        True \vee b_{ii}^{k-1} \quad \text{if converged}, \\ 
        False \vee b_{ii}^{k-1} \quad \text{else}.
    \end{cases}
\end{equation}
The remaining boolean values are updated using logical consensus
\begin{equation}
    b_{ij}^k = \bigvee_{h\in\{i\}\cup\mathcal{N}_i} b_{hj}^{k-1} \ \forall \ j \neq i \in \mathcal{V},
\end{equation}
the distance values are updated using min consensus
\begin{equation}
    d_{ij}^k = \min(d_{ij}^{k-1},\min_{h\in\mathcal{N}_i}d_{hj}^{k-1}+1), \ \forall \ j\neq i \in \mathcal{V},
\end{equation}
and the switching time step is also updated using min consensus
\begin{equation}
    s_i^k = \min_{j\in\{i\}\cup\mathcal{N}_i}s_j^{k-1}.
\end{equation}
Under the assumption that one message pass takes a single time step and that information spreads in a wavefront, if robot $i$ knows that all robots are ready, it also knows that the highest number of time steps it takes for the other robots to know this is equal to the largest distance in the network. Therefore, the latest time step that all robots are aware that all robots are ready to transition is equal to the current time step plus the largest element of the distance vector $\bm{d_i^k}$. Using this, robot $i$ sets its transition time to
\begin{equation}
    s_i^k = \min(s_i^{k},k + \max(\bm{d_i^k}))), \ if \ \bigwedge_{j\in\mathcal{V}}b_{ij}^k = True.
\end{equation}

This approach can be interpreted as a dynamic leader election. When robots are ready, their boolean value will propagate through the graph like a wavefront. The robot that is first to be aware that all robots have converged updates the switching time step and propagates it back to all the robots in a wavefront.
\subsection{Adjacency Estimation Convergence Criteria}
The condition for the adjacency matrix estimation having converged on each robot is the following
\begin{equation}
    est. \ cvg. = True \Leftrightarrow \max_{(j,l)\in\mathcal{V}\times\mathcal{V}} |\hat{a}_{jl,i}^k - \hat{a}_{jl,i}^{k-1}| \leq \zeta,
\end{equation}
i.e., when all the entries of the adjacency matrix have converged.
\subsection{Optimisation Convergence Criteria}
The optimisation on each robot is considered converged when the absolute normalised change in the penalty multiplier is within some bound $\eta$
\begin{equation}
    opt. \ cvg. = True \Leftrightarrow \max_{j\in\mathcal{N}_i} \left | \frac{\bm{\mu_i^k}}{\bm{\mu_i^{k-1}}} -1 \right | \leq \eta,
\end{equation}
i.e., when the trading variables have converged.

\section{Algorithm}
\begin{figure}[ht]
    \centering
    \begin{tikzpicture}[node distance=1.5cm, every text node part/.style={align=center}]
        \node (start) [startstop] {Set Reference to Current Position};
        \node (mtr) [process, below of=start] {Move toward Reference};
        \node (uae) [process, below of=mtr] {Update Adjacency Matrix \\ Estimate \& Convergence Variables};
        \node (cvg1) [decision, below=0.75cm of uae] {$k = s_i^k$};
        \node (exf) [process, below=0.75cm of cvg1] {Exchange Adjacency Matrix \\ Estimate \& Convergence Variables};
        \node (cfe) [process, below of=exf] {Reinitialise Convergence Variables \& \\ Calculate Fiedler Estimate};
        \node (sop) [process, below of=cfe] {Solve Optimisation Problem \& \\ Update Convergence Variables};
        \node (cvg2) [decision, below=0.75cm of sop] {$k = s_i^k$};
        \node (ext) [process, below=0.75cm of cvg2] {Exchange Trading \\ \& Convergence Variables};
        \node (sop2) [process, below of=ext] {Average trading variables \& \\ solve optimisation problem wo. trading};
        \node (upr) [process, below of=sop2] {Reinitialise Optimisation and Convergence \\ Variables \& Update Reference};

        \node[draw=none] (mtr_r) [right=1 cm of mtr.west] {};

        \draw[->] (start.south) -- (mtr.north);
        \draw[->] (mtr.south) -- (uae.north);
        \draw[->] (uae.south) -- (cvg1.north);
        \draw[->] (cvg1.south) -- node[right] {False} (exf.north);
        \draw[->] (cfe.south) -- (sop.north);
        \draw[->] (sop.south) -- (cvg2.north);
        \draw[->] (cvg2.south) -- node[right] {False} (ext.north);
        \draw[->] (ext.south) -- (sop2.north);
        \draw[->] (sop2.south) -- (upr.north);
        \draw[->] (cvg1.west) -- node[above] {True} ++(-2cm,0) |- (cfe.west);
        \draw[->] (exf.east) -- ++(1cm,0) |- (uae.east);
        \draw[->] (cvg2.west) -- node[above] {True} ++(-2cm,0) |- (sop2.west);
        \draw[->] (ext.east) -- ++(1cm,0) |- (sop.east);
        \draw[->] (upr.south) -- ++(0,-0.5cm) -- ++(-4.5cm,0) |- (mtr.west);
    \end{tikzpicture}
    \caption{D-MR-CaTP algorithm structure.}
    \label{fig:algorithm}
\end{figure}
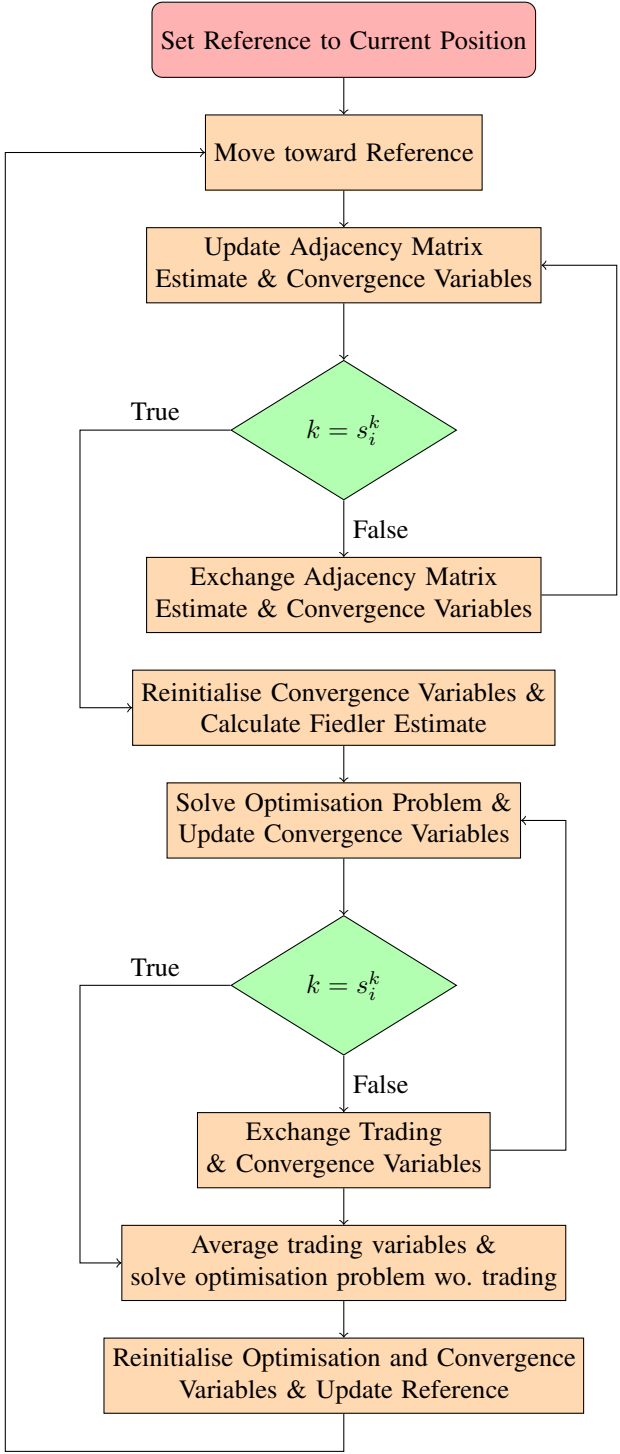

The distributed MR-CaTP (D-MR-CaTP) algorithm runs synchronously across all robots. The algorithm runs as follows: The robots move towards their position references using their onboard controllers for a fixed number of time steps. The robots then halt and starts estimating the adjacency matrix until convergence. After having estimated the adjacency matrix, the robots reinitialise their convergence variables and distributively compute the optimal trajectories until convergence. Once the optimisers have converged, the robots average their trading variables, perform the optimisation once more without trading, reinitialise their convergence variables, and update their position reference. This entire procedure is then repeated indefinitely or until termination. The algorithm is summarised in \cref{fig:algorithm}

\section{Simulation Results}
To verify the efficacy of the method, it is tested in an inspection task with $N=10$ planar robots, where a subset of the robots $\mathcal{I}\subseteq\mathcal{V}$ are assigned to inspect POIs and the remaining robots support the inspection robots. The cost function for the robots are
\begin{equation}
    f_i(\bm{u_i^k}) = \begin{cases}
        \dfrac{1}{2} ||\bm{p_{poi}} - (\bm{p_i^k} + \bm{u_i^k})||_2^2 + h||\bm{u_i^k}||_2^2, \ \text{if} \ i \in \mathcal{I},
        \\
        h||\bm{u_i^k}||_2^2, \ \text{else},
    \end{cases}
\end{equation}
where the inspection robots cost function is the squared distance to their respective POIs and a cost on movement, and the support robots only have a cost a cost on movement. Furthermore, the constraint in \eqref{eq:vel_const} uses the infinity norm.
\begin{figure}[ht]
    \centering
    \includegraphics[width=\linewidth]{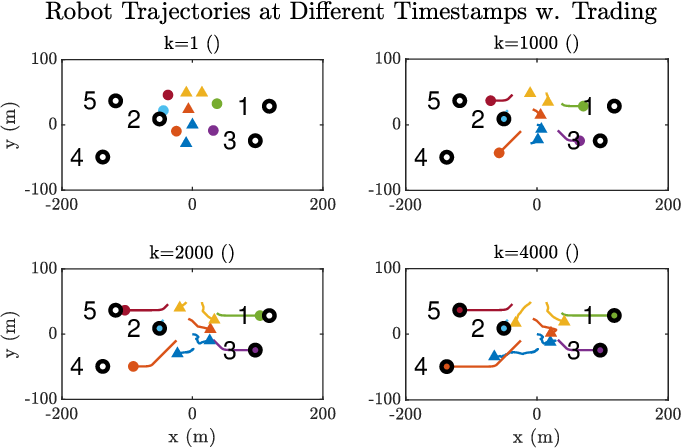}
    \caption{Snapshots of simulation at four different iterations of D-MR-CaTP algorithm with 5 inspection points and $N=10$ robots. The circles indicate inspection robots while the triangles indicate support robots. The POIs are indicated by the black circles with the numbering of the POI being adjacent.}
    \label{fig:dist_poi_pos}
\end{figure}
\begin{figure}[ht]
    \centering
    \includegraphics[width=\linewidth]{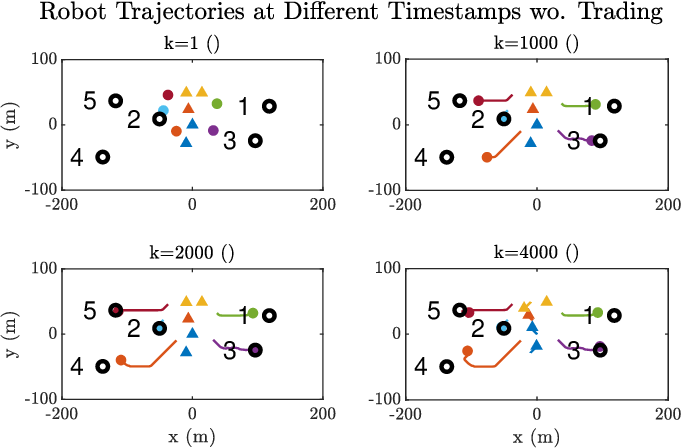}
    \caption{Snapshots of simulation  at four different iterations of D-MR-CaTP algorithm with 5 inspection points and $N=10$ robots. The circles indicate inspection robots while the triangles indicate support robots. The POIs are indicated by the black circles with the numbering of the POI being adjacent.}
    \label{fig:dist_poi_pos_wo_trading}
\end{figure}
In \cref{fig:dist_poi_pos} and \cref{fig:dist_poi_pos_wo_trading} the trajectories of the robots, when the robots are allowed and not allowed to trade budget, can be seen. When the robots are allowed to trade budget, the inspection robots effectively "pull" the support robots since they request additional budget, which the support robots can only provide by moving to increase the Fiedler value. This happens until the inspection robots reach the POIs, at which point the inspection robots no longer require additional budget and the robots terminate their movement. When the robots are not allowed to trade budget, the inspection robots move towards the POIs, while the support robots stand still, until they have no more communication budget. The support robots briefly move to ensure that the Fiedler value remains above the lower bound $\underline{\lambda}_2$.
\begin{figure}[ht]
    \centering
    \includegraphics[width=\linewidth]{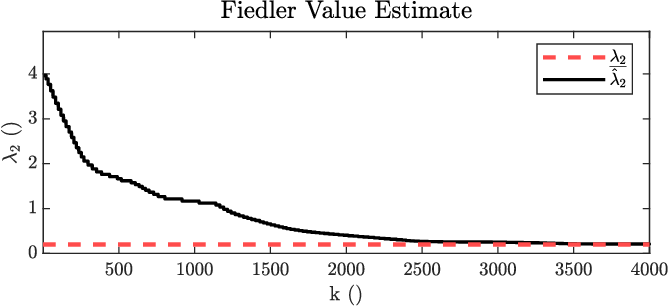}
    \caption{Fiedler value estimate for POI inspection task simulation. The Fiedler value estimate remains above the lower bound for all time.}
    \label{fig:dist_poi_fiedler}
\end{figure}
As can be seen in \cref{fig:dist_poi_fiedler}, the Fiedler value remains above the lower bound $\underline{\lambda}_2$ for the duration of the simulation. Unlike in \cite{Mikkelsen2024OptimalValue} where the support robots move to maximise the Fiedler value, in this simulation the Fiedler value terminates close to the lower bound as there is no incentive for the robots to maximise it.
\begin{figure}[ht]
    \centering
    \includegraphics[width=\linewidth]{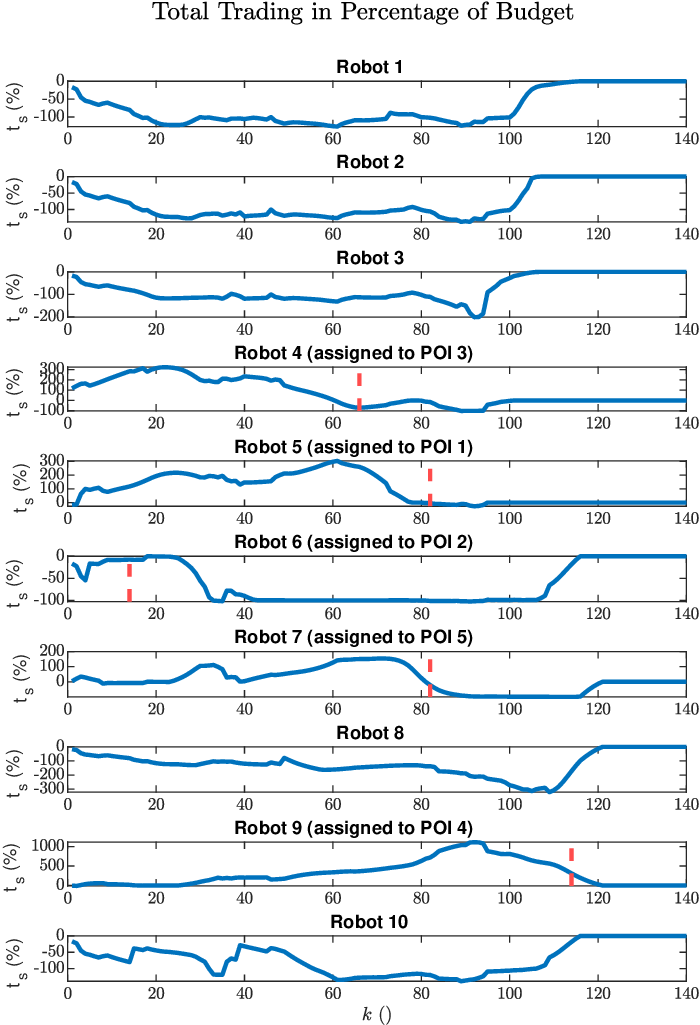}
    \caption{Sum of all trading of each robot as a percentage of its current budget. The red dashed vertical lines indicate when the robot is within $1$ m of its assigned POI. It should be noted that since the robots do not perform optimisation at every iteration, the number of iteration rounds are less than the total number of iterations of the simulation.}
    \label{fig:dist_trading_percent}
\end{figure}
In \cref{fig:dist_trading_percent}, the sum of all the trading variables for each robot as a percentage of their current budget can be seen
\begin{equation}
    t_{s,i}^k = \frac{\sum_{j\in\mathcal{N}_i} t_{ij}}{(\hat{\lambda}_{2,i}^k - \underline{\lambda}_2)/N}. 
\end{equation}
It can be noted that the net trading of the support robots 1, 2, 3, 8, and 10 are all negative since they have no incentive to request additional budget. The net trading of the inspection robots 4, 5, 7, and 9 are for the most part positive, since they require additional budget to move to the POIs. However, they do for some duration of the time sell budget, which is due to them not reaching their POIs simultaneously, and therefore the robots that reach their POIs first can trade away their budget to other robots that need it. Interestingly, robot 6 which is assigned to POI 2 has net negative trading for the entire duration of the simulation. In \cref{fig:dist_poi_pos} it can be seen that this is due to the robots initial position being in close vicinity to its assigned POI.
\begin{figure}[ht]
    \centering
    \includegraphics[width=\linewidth]{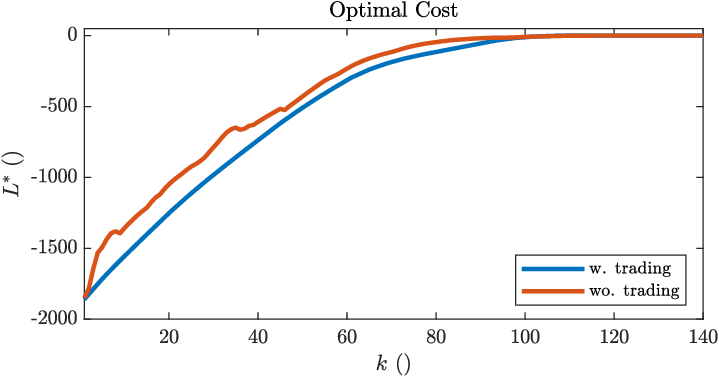}
    \caption{Optimal cost for distributed optimisation with and without trading for each optimisation round. It should be noted that since the robots do not perform optimisation at every iteration, the number of iteration rounds are less than the total number of iterations of the simulation.}
    \label{fig:dist_opt_cost}
\end{figure}
In \cref{fig:dist_opt_cost}, the optimal cost for each round of the distributed optimisation problem can be seen. Furthermore, for each completed round of the distributed optimisation, the optimal cost of the distributed optimisation without trading can be seen\footnote{It should be noted that this is not the optimal cost for the simulation seen in \cref{fig:dist_poi_pos_wo_trading}}. As can be seen, the cost for the distributed optimisation with trading is noticeably lower than without trading, demonstrating a clear benefit in allowing the robots to trade budget.
\begin{table}[ht]
    \centering
    \caption{Statistics of running time and number of steps for each iteration of the state machine in \cref{fig:algorithm}.}
    \begin{tabular}{c|c|c|c|c}
                          & \textbf{min} & \textbf{median} & \textbf{mean} & \textbf{max} \\
        \hline \hline
        time & 7.4e-3 & 1.9e-2 & 2.3e-2 & 1.2e-1 \\
        \hline
        steps & 13 & 26 & 28 & 112
    \end{tabular}
    \label{tab:stat_table}
\end{table}
Lastly, in \cref{tab:stat_table} statistics on the running time and number of steps required to make it through the state machine presented in \cref{fig:algorithm} can be seen. When comparing it with the statistics for the centralised MR-CaTP algorithm in \cite{Mikkelsen2024OptimalValue}, it can be seen that on average the algorithm is 8 times slower. This does not take into consideration the time it takes for message passing in the algorithm though, which can significantly affect the statistics. It should be noted however, that the statistics in \cite{Mikkelsen2024OptimalValue} also does not take into account the time it takes to transmit the optimal trajectories to the robots in the network.

\section{Discussion}


The following is a discussion on the proposed method and its limitations.
\subsection{Comparison with Centralised Optimisation Approach}
A quantitative comparison between the distributed optimisation approach presented in this paper and the centralised optimisation approach presented in \cite{Mikkelsen2024OptimalValue} is not possible beyond comparing the time and number of steps to compute a solution, as the cost functions of the two methods are different. However, a qualitative comparison can be made. The main difference between the two methods is their underlying mechanics. While the centralised method relied on moving the support robots along the gradient of the Fiedler value, the method presented in this paper relies on the inspection robots effectively "pulling" the support robots by requiring more budget than they can provide at a given time. The outcome of this difference is that while in the centralised approach, when the inspection robots reach their POIs the support robots keep moving until they reach a stationary point with regard to the Fiedler value gradient (see Fig.~5 in \cite{Mikkelsen2024OptimalValue}), in the distributed approach once the inspection robots reach their POIs they will no longer be incentivised to trade budget with the support robots and the support robots will therefore cease moving. This results in the Fiedler value being stationary once the POIs have been reached, as can be seen in \cref{fig:dist_poi_fiedler}.
\subsection{Separation of Communication Constraint}
Instead of separating the communication constraint, it is possible to solve the optimisation problem distributively by having each robot solve the optimisation problem over the full problem variable set and using, e.g., C-ADMM to reach consensus on the optimal solution \cite{Boyd2010DistributedMultipliers}. This poses the challenge of scalability, since the complexity of the optimisation problem grows exponentially with the number of decision variables, which is linear with the number of robots. Using the approximate separation of the communication constraint, the complexity of the optimisation problem for each robot is reduced, since it only needs to find the optimal solution for its own trajectory as well as the optimal trading variables, which are linear with the number of communication neighbours. For a system with $N$ agents, in the worst case all robots communicate with each other, i.e., there are $M(N-1)$ trading variables, where $M$ is the length of the prediction horizon. If each robot had to optimise for all robots, there would be $nM(N-1)$ decision variables added to the decision variables for its own trajectory, where $n$ is the number of states. Clearly, this means that in the worst case scenario, it is still beneficial for the robots to use the approximate separation of the communication constraint.
\subsection{Dynamic Feasibility}
The presented method assumes that the robots are holonomic and that they can track the desired trajectory. For ground robots with omnidirectional drive and multi-rotor UAVs this assumption can be valid, but still limiting. Including the dynamics of the robots in the optimisation problem is possible, but will further increase the complexity. Furthermore, it can risk over constraining the optimisation problem such that there are no feasible solutions.
\subsection{Complexity}
Since the robots distributively estimate the adjacency matrix, the larger the diameter of the communication network graph is, the longer it takes for the adjacency matrix estimate to converge, making the estimation slower. Conversely, if the robots have many connections, making the adjacency estimate converge quickly, the complexity of the optimisation problem on each robot grows, making the optimisation slower.
\subsection{Convergence Consensus}
The convergence consensus method allows the robots to distributively reach agreement on when to transition between states. However, to do this each robot needs to store a boolean vector with the same size as the number of robots. If one of the robots suddenly fails and is unable to participate in the consensus, it will halt all the other robots. To mitigate this, the addition of failure detection across the network would be prudent, allowing the robots to exclude robots that have failed from the formation. This is not within the scope of this paper, and has therefore not been considered.

\section{Conclusions}
In this paper, a distributed multi-robot communication-aware trajectory planning method was presented. The method uses an economic interpretation to make an approximate separation of a non-separable linear inequality constraint. The optimisation problem was solved distributively using a dual-ascent approach to enforce consensus between the robots on the exchange of communication budget. The efficacy of the method was demonstrated through simulation in an inspection task.
\appendix
\subsection{Fiedler Value Derivative}\label{sec:fiedler_derivative}
The derivative of the Fiedler value with regard to the change in position can be found from \cite{Mikkelsen2024OptimalValue} as
\begin{equation}\label{eq:fiedler_gradient}
    \begin{split}
        \bm{m_i^{k}} = \left[ \bm{v_2^{k\top}}\frac{d\mathbf{L}}{dp_{i,1}}\bigg\rvert_{p_{i,1}=p_{i,1}^k}\bm{v_2^k},\dots \right. \\ \left. , \bm{v_2^{k\top}}\frac{d\mathbf{L}}{dp_{i,n}}\bigg\rvert_{p_{i,n}=p_{i,n}^k}\bm{v_2^k} \right]^\top \in \R^{n}.
    \end{split}
\end{equation}
The derivative of the Laplacian with respect to $p_{i,r}$ is:
\begin{equation}
    \frac{d\mathbf{L}}{dp_{i,r}} = \frac{d\mathbf{D}}{dp_{i,r}} - \frac{d\mathbf{A}}{dp_{i,r}},
\end{equation}
where the derivatives of the adjacency and degree matrices are:
\begin{equation}
    \frac{da_{ii}}{dp_{i,r}} = 0, \ \forall \ i \in \mathcal{V}, \ r \in \N_n,
\end{equation}
\begin{equation}
    \begin{split}
    \frac{da_{ij}}{dp_{i,r}} = \frac{da_{ji}}{dp_{i,r}} = -\alpha(1 - a_{ij}) a_{ij} \frac{p_{i,r} - p_{j,r}}{||\bm{p_i} - \bm{p_j}||_2}, \dots \\ \forall j \in \mathcal{N}_i, \ r \in \N_n,
    \end{split}
\end{equation}
\begin{equation}
    \frac{da_{ij}}{dp_{i,r}} = 0 \ \forall \ j \notin \mathcal{N}_i,
\end{equation}
\begin{equation}
    \frac{da_{jh}}{dp_{i,r}} = 0, \ \forall \ j,h \neq i \in \mathcal{V}, \ r \in \N_n,
\end{equation}
\begin{equation}
    \frac{d\mathbf{D}}{dp_{i,r}} = \text{diag}\left(\frac{d\mathbf{A}}{dp_{i,r}} \bm{1}\right),
\end{equation}
where $\mathcal{N}_i$ is the set of robots with which robot $i$ has a communication link.
\printbibliography

\end{document}